# INTEGRATING NATURAL LANGUAGE PROCESSING TECHNIQUES OF TEXT MINING INTO FINANCIAL SYSTEM: APPLICATIONS AND LIMITATIONS

D. Millo    B. Vika    N. Baci

*Statistics and Applied Informatics Department, Faculty of Economy, University of Tirana, Tirana, Albania*
*denisa.millo@unitir.edu.al, blerina.vika@unitir.edu.al, nevila.baci@unitir.edu.al*

**Abstract-** The financial sector, a pivotal force in economic development, increasingly uses the intelligent technologies such as natural language processing to enhance data processing and insight extraction. This research paper through a review process of the time span of 2018-2023 explores the use of text mining as natural language processing techniques in various components of the financial system including asset pricing, corporate finance, derivatives, risk management, and public finance and highlights the need to address the specific problems in the discussion section. We notice that most of the research materials combined probabilistic with vector-space models, and text-data with numerical ones. The most used technique regarding information processing is the information classification technique and the most used algorithms include the long-short term memory and bidirectional encoder models. The research noticed that new specific algorithms are developed and the focus of the financial system is mainly on asset pricing component. The research also proposes a path from engineering perspective for researchers who need to analyze financial text. The challenges regarding text mining perspective such as data quality, context-adaption and model interpretability need to be solved so to integrate advanced natural language processing models and techniques in enhancing financial analysis and prediction.

## 1. INTRODUCTION

The financial sector is an important and a driving force of the worldwide economic development and it is having a rapid development due to the use of the latest intelligent technologies such as Natural Language Processing (NLP). The period of 2018-2023 covered in this research paper implies an important one, where NLPs have seen an exponential growth with the Large Language Models (LLMs) projects like ChatGPTs, Gemini, Bard and Bing exploration and models deeper specifications.

For the scope of this research paper, we refer to the Finance field as the Financial System (FS) as it encompasses various components such as; public finance, international finance, corporate finance, derivatives, risk management, portfolio theory, asset pricing (AP), and financial economics [1].

A detailed view of the search methods and machine learning tools in language processing, including classifiers and sequence models, techniques like decision trees (DT), support vector machines (SVM), and cross-validation is essential for evaluating NLP systems [2]. We will give examples of the use of NLP techniques in the next sections in this research, enhancing their understanding by delivering the research questions as following:

Q1: What are the NLP models and techniques used for analyzing FS components in the time span of 2018-2023?
Q2: What are the challenges and limitations in the applications of NLP to the FS regarding text-mining applications on the engineering perspective of data processing?

## 2. THE APPLICATIONS OF NLP MODELS AND TECHNIQUES IN FINANCIAL SYSTEM

Going deeper into example cases, we discover an ongoing trend that is growing fast. To analyze the subject-matter we go through a time span of 2018-2023.

### 2.1. Materials and Methods

The path used to answer to Q1 is built on the methodology of filtering in the Google Scholar search engine. There are included citations and patents with the query as follows: (NLP OR "Natural Language Processing") AND "*FS element mentioned in chapter 1"AND text mining AND forecasting AND prediction AND models. To give answer to our research questions we explored around 250 scientific materials, from which 35 of them books and the rest papers and articles. After selecting the scientific materials filtered, we reached to a total of around one hundred and nine (109) papers, articles, books and similar. For the scope of this research paper, we classified and analyzed around sixty (60) study materials.

**2.2. Results**

Figure 1 shows that the focus of most studies relies on AP, (especially on stock prediction), while having the corporate finances as the second most studied component of FS.

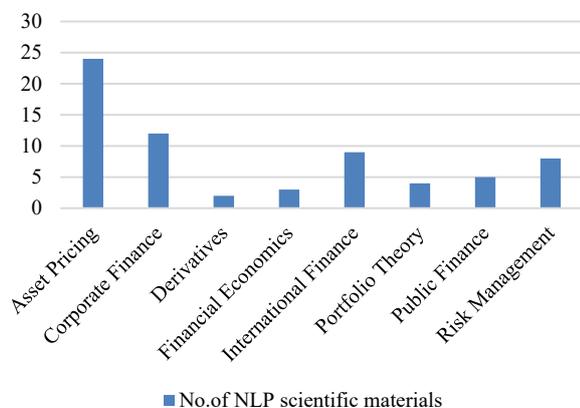

Figure 1. Histogram of scientific materials extracted and classified according to Financial System Components for the years 2018-2023

From the actual database of scientific materials, we also noticed that most of NLP models were a combination of probabilistic and vector-based ones. Whereas, regarding the classification on NLP techniques evidenced, we can refer to Figure 2. As we can see, the information classification techniques and the combination of the all three ones, has most of the works used during the time span study.

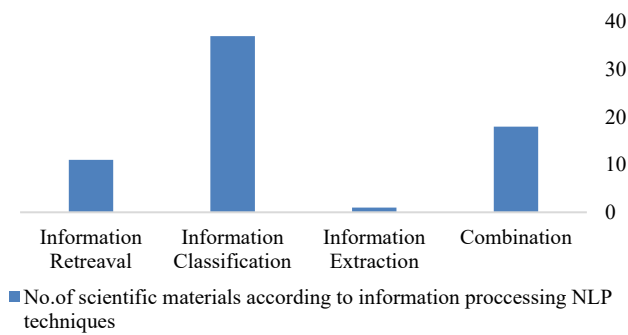

Figure 2. Histogram of scientific materials extracted and classified according to information processing NLP techniques, years 2018-2023,

We noticed a growing number of articles which combined both probabilistic with VSM, and also combined text-data with numerical ones. A systematic and critical review of deep learning in financial economics is provided by using hybrid variants of methods [3]. The work [4] forecasted economic trends using sentiment analysis, AutoRegressive with eXogenous inputs (ARX) and Lasso on data from articles from six newspapers showing varying levels of accuracy. The Numeric Hierarchical Transformer Model (NumHTML) for financial forecasting using earning conference calls data was introduced in the year of 2022 [5]. As these studies remain near the financial economics area, they also might have infusions of other topics, showing at least, a low interest on the matter.

A textual analysis of China's financial markets is conducted with sentiment analysis [6]. The work of [30] used transformers for sentiment analysis on Environmental, Social, and Governance (ESG) news. Datasets on the international finance FS component are of various data sources, as they help to create an international point of view on the topic.

As for the portfolio theory application of NLPs, we can elaborate works that underpin the interconnection of the fields of FS and NLP. An important work [7] applies financial Bidirectional Encoder Representations from Transformers (FinBERT), Monte Carlo, and Black and Litterman model for portfolio optimization on Milano Finanza data. Whereas, [28] investigated the importance of financial sentiment in portfolio management using RF, Multi-Layer Perceptron (MLP), and LSTM on data from Quandl and StockFluence APIs. Here we have to mention even the use of datasets from other sectors for training models, which can be sub-optimal, highlighting the need for more sector-specific annotated datasets [8,9,10]. The diverse number of techniques used, show of a versatile sub-field, where data and information are not absent, but needs more attention in the future, for taking proper financial decisions and predictions to the investors.

The public finance component is elaborated in some works below though, there is a need to be further investigated. Work of [11] applied machine learning to budget speech statements. The article of [12] examined tax compliance behavior using Bidirectional Encoder Representations from Transformers (BERT) on survey responses. A bank financial risk prediction model is developed using big data from commercial banks, employing Least Absolute Shrinkage and Selection Operator (Lasso) and SVM [29]. The SVM techniques [13] and hybrid deep learning models [14] show their global use in different fields, we can mention the healthcare one. Whereas the finance field is evolving, the need for intelligent methods is rising, as it is a crucial point when taking financial decisions, and is applied thoroughly in every other component presented above.

Through our work we saw that the stock predictions remain the most studied topic, in the AP component of the FS, and that the AP component of the FS is the most studied topic compared to other components. There might be different reasons, but what we can deduce is that there are more open databases that allow the latter topic to be studied and also, it is a very fast needed information. An investor has to have an accurate and timely information on stocks or in AP, in order to take proper decision. There is evidenced a variety of datasets utilized, including financial news databases. stock returns data, corporate financial reports (e.g., 10-K filings), and social media data (e.g., StockTwits, Weibo).

The most used common specific techniques, as presented in Figure 3. include BERT types, Long Short-Term Memory (LSTM) and Logistic Regression (LR), whereas Lasso, Bag of Words (Bow), NEUS, K-Nearest Neighbors (KNN), Deep Neural Network (DNN), Deep Learning (DL), Generalized Autoregressive Conditional Heteroskedasticity (GARCH) and Autoregressive

Integrated Moving Average (ARIMA) are to be mentioned. Other types of techniques, which are a combination of above and new techniques proposed e.g. NumHTML, AdaBoost, compose a share of 47 % in overall scientific material of 60 papers used for the purpose of this section. The frequently used evaluation parameters were Accuracy, F1-score, Precision, Recall, and R2.

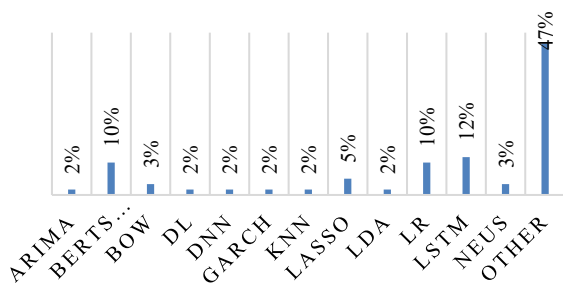

Figure 3. Histogram of scientific materials extracted and classified according to specific NLP techniques, years 2018-2023, Source: Authors

## 3. DISCUSSION

This research highlighted several limitations on the text engineering perspective in the application of NLP to FSs in order to answer to Q2. It leverages the past works' [15] limitations and challenges (restriction to confidential data, absence of well-defined financial lexicon lists, lack of dynamic texts analysis models, need of club inter-domain results, highly un-structured and redundant data, sarcasm and vernacular language) are not fully appointed yet. Let us point out the evolution of the challenges from the time of the study's [15] analysis to the end of the year 2023. A significant issue is the data shift between training corpora and real estate headlines, leading to poor performance in event extraction [16]. Traditional models like the Fama-French 5-factor model (FF5) face overfitting problems, highlighting superiority of newer models like Generalized News-based Sentiment Analysis (GNUS) [17].

Whereas even the latter like deep learning models encounter challenges with Non-Independent and Identically Distributed (non-IID) samples, time-varying distributions, and low signal-to-noise ratios, complicating stock return predictions such as in [18]. Refs. [17, 18] indicate a partially addressed issue of Gupta, et al. [15] challenges to the lack of dynamic text analysis models. There is a limited interpretability of advanced models like BERT and Word2Vec that restricts their application, particularly when not trained on domain-specific texts [19, 20]. The sentiment analysis models often fail to capture financial context adequately due to reliance on general lexicons [21], a challenge not yet solved since the research paper [15] reference on absence of well-defined financial lexicon lists.

Further, data quality and causality issues persist, making it difficult to establish reliable links between sentiment indicators and outcomes [22]. The overreliance on numerical data in models overlooks qualitative factors like managerial experience, which could enhance predictions [23]. The matter of scalability and adaptability of models to real-time data remain challenging, particularly in evolving financial markets [24]. Even the expert annotations based on prior beliefs can be incorrect, emphasizing need for context-based annotations [25]. The infancy of text analysis in finance requires comprehensive research to fully understand its potential [4].

Additionally, the use of datasets from other sectors for training models can be sub-optimal, highlighting the need for more sector-specific annotated datasets [10]. Future research should take into consideration market frictions and transaction costs in trading strategies [26]. While some [15] challenges are mentioned above, the other ones such restriction to confidential data, need of club inter-domain results, highly un-structured and redundant data, sarcasm and vernacular language and the new ones we presented above resonate a need to address them as quickly as possible. Lastly, we can state that advanced NLP models show promise but require further investigation to address their current limitations and optimize their use in financial predictions [27].

Table 1. Challenges and Limitations in the application of NLP to FS, the x symbol means that the problem persists and needs to be addressed

| Challenges and Limitations in the application of NLP to FSs regarding text mining engineering perspective | Gupta, et al. [15] | This research |
|---|---|---|
| restriction to confidential data | x | solved |
| absence of well-defined financial lexicon lists | x | x |
| lack of dynamic texts analysis models | x | partially resolved |
| need of club inter-domain results | x | x |
| highly un-structured and redundant data | x | x |
| sarcasm and vernacular language | x | x |
| data shift between training corpora and real-world corpora | | x |
| overfitting problems of traditional model | | x |
| limited interpretability of advanced models | | x |
| time-varying distributions | | x |
| low signal-to-noise ratios | | x |
| fail to capture the financial context adequately due to reliance on general lexicons | | x |
| data quality and causality issues | | x |
| matter of scalability and adaptability of models to real-time data | | x |
| incorrect expert annotations based on prior beliefs | | x |

As per conclusion of this chapter, we propose a path to be followed when having to do research on the finance field combined with NLP models and techniques of text mining. Figures 4 and 5 explain visually the steps as per follows.

After determining the where we will delve in while following the Figure 4, we follow the text analyzing steps mentioned in Jurafsky and Martin 2020 [2]. Figure 5 explains in detail the broader panorama of analyzing financial text in the language context and the specific limitations encountered from our work in the research papers analyzed. In the right side of each step, there are mentioned the most encountered limitations. These limitations should be taken into account when dealing with problems of the same field as this research paper handles.

Financial texts remain difficult to understand as they have a specific domain language. The models and limitations discussed in this research paper tend to make easier the first stage of a research, the state of the art. After choosing the FS component field, NLP model and techniques, the language analysis and having into account the proper limitations as expressed in Figure 5, the data preparation should begin.

**Asset Pricing** — Information Classification, LSTM
• S&P, Bloomberg, Dow Jones etc.

**Corporate finance** — Combination, Other
• 10-K and 10-Q reports, etc.

**Derivatives** — Combination, Other
• Thomson Reuters etc.

**Risk Managment** — Combination, Other
• Industries annual reports etc.

**International Finance** — Information Classification, Other
• News articles, International Banks data etc.

**Portfolio Theory** — Information Classification, Other
• Wall Street, Dow Jones etc.

**Financial Economics** — Information Classification, Other
• World Bank, Yahoo Finance etc.

**Public Finance** — Information Retrieval, BERTs
• Budget speeches etc.

Figure 4. The first step is to determine the FS component to work on with and the NLP technique regarding information processing and model algorithms. In the left side FS components, in the right side the most used NLP techniques for each component. In the bottom the most probable data source. "Other" and "Combination" have the same meaning as mentioned in Figures 1 and 2. In this Figure there are mentioned the most used NLP techniques and models

## 4. CONCLUSIONS

Despite the limitations, NLP holds substantial promise for enhancing the FS. Their mixed applications can significantly improve the accuracy of predictions, such as stock prices and asset returns, by taking large amounts of textual data from news articles, social media, and financial reports. Advanced models like BERT and LSTM, and their modified models, have demonstrated superior performance in capturing the nuances of financial language and predicting market movements. Integrating deep learning techniques into financial models not only enhances prediction accuracy but also helps uncover underlying economic mechanisms, offering deeper empirical insights into AP. Furthermore, NLP can facilitate better sentiment analysis, crucial for gauging market sentiment and making informed investment decisions.

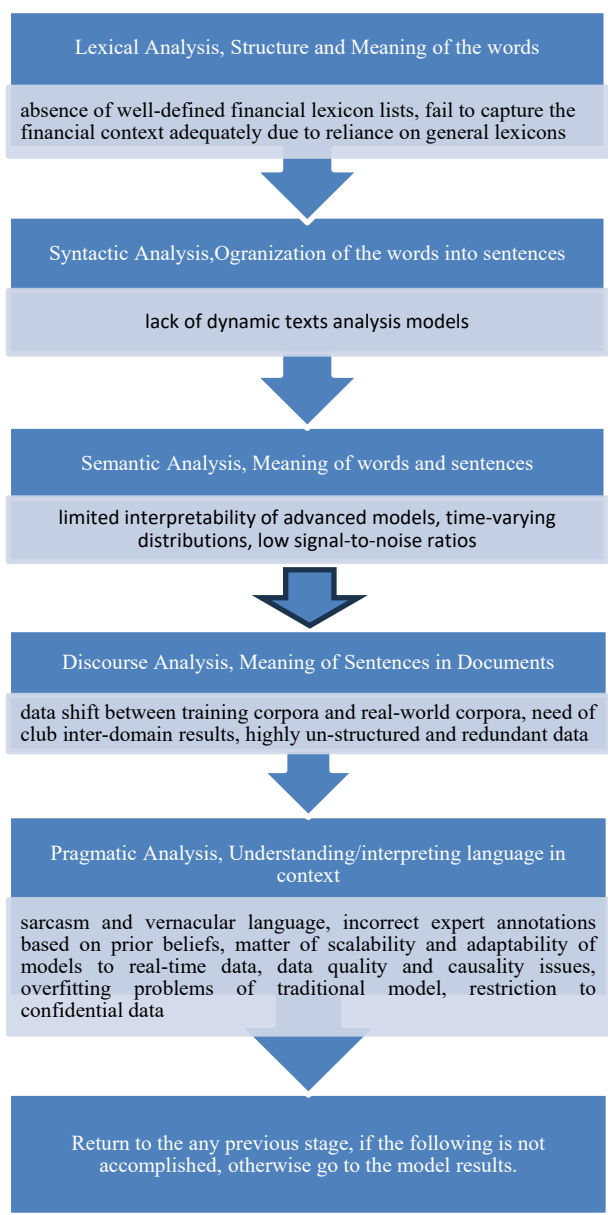

Figure 5. Second step is to determine the problem on language terms and in which phase of language analysis it will go on to. For each double headed sub diagram, there is the step analysis on top and limitations in the bottom. Source: Martin and Jura sky (2020) with the limitations from the authors

Advancements in transfer learning and domain adaptation are promising for addressing data shift issues, thereby improving model robustness and generalization. As the field progresses, developing more interpretable AI models will enhance transparency and trust, making NLP an indispensable tool for financial analysts and researchers.

The FS components of AP and corporate finance remain amongst most elaborated with NLP techniques, while the other components have a small number of studies. Amongst NLP models, the most used remain the probabilistic and VSM models, whereas the most used techniques remain the information classification and hybrid techniques. Overall, while advanced NLP and deep learning models show great promise for financial applications, ongoing research must address these limitations to enhance their accuracy, interpretability, and applicability. Future research should focus on refining models to account for local market dynamics, improving data quality and annotation methods, and developing more sophisticated algorithms capable of handling the complex and dynamic nature of financial data.

## ACKNOWLEDGEMENTS

The author(s) appreciate the assistance of the finance student Savina Shtaka for supporting in material search.

## REFERENCES

[1] T. Hens, M.O. Rieger, "Financial Economics", Springer Texts in Business and Economics, Springer Berlin Heidelberg, pp. 5-6, Heidelberg, Berlin, Germany, 2016.
[2] J. Martin, D. Jurafsky, "Speech and Language Processing an Introduction to Natural Language Processing", Computational Linguistics, and Speech Recognition, 3rd ed., pp. 3-263, 2020.
[3] Y. Zheng, Z. Xu, A. Xiao, "Deep Learning in Economics: A Systematic and Critical Review", Arif. Intell. Rev., Vol. 56, pp. 9497-9539, September 2023.
[4] L. Barbaglia, S. Consoli, S. Manzan, "Forecasting with Economic News", Journal of Business and Economic Statistics, Vol. 41, pp. 708-719, July 2023.
[5] L. Yang, J. Li, R. Dong, Y. Zhang, B. Smyth, "NumHTML: NumericOriented Hierarchical Transformer Model for Multi-Task Financial Forecasting", AAAI, Vol. 36, pp. 11604-11612, June 2022.
[6] A. Huang, W. Wu, T. Yu, "Textual Analysis for China's Financial Markets: a Review and Discussion", China Finance Review International, Vol. 10, No. 1, pp. 1-15, 2020.
[7] F. Colasanto, L. Grilli, D. Santoro, G. Villani, "BERT's Sentiment Score for Portfolio Optimization: A Fine-Tuned View in Black and Litterman Model", Neural Comput. and Applic., Vol. 34, pp. 17507-17521, October 2022.
[8] M. Wujec, "Analysis of the Financial Information Contained in the Texts of Current Reports: A Deep Learning Approach", JRFM, Vol. 14, p. 582, December 2021.
[9] C. Chen, S. Xiao, B. Zhao, "Machine Learning Meets the Journal of Public Budgeting and Finance: Topics and Trends Over 40 Years", Public Budgeting and Finance, Vol. 43, pp. 3-23, December 2023.
[10] T.I. Theodorou, A. Zamichos, M. Skoumperdis, A. Kougioumtzidou, K. Tsolaki, D. Papadopoulos, T. Patsios, G. Papanikolaou, A. Konstantinidis, A. Drosou, D. Tzovaras, "An AI-Enabled Stock Prediction Platform Combining News and Social Sensing with Financial Statements", Future Internet, Vol. 13, p. 138, May 2021.
[11] C. Vuppalapati, A. Ilapakurti, S. Vissapragada, V. Mamaidi, S. Kedari, R. Vuppalapati, S. Kedari, J. Vuppalapati, "Application of Machine Learning and Government Finance Statistics for Macroeconomic Signal Mining to Analyze Recessionary Trends and Score Policy Effectiveness", IEEE International Conference on Big Data (Big Data), pp. 3274-3283, Orlando, FL, USA, December 2021.
[12] I. Florina, C. Stefana, C. Codruta, "Is Trust a Valid Indicator of Tax Compliance Behavior? A Study on Taxpayers' Public Perception Using Sentiment Analysis Tools", A.M. Dima, M. Kelemen, (Eds.), "Digitalization and Big Data for Resilience and Economic Intelligence", Springer Proceedings in Business and Economics. Springer, Cham, Switzerland, 2022.
[13] V. Jain, B. Jha, S. Joshi, S. Miglani, A. Singal, S. Babbar, M. Demirci, M.C. Taplamacioglu, "Human Disease Detection Using Artificial Intelligence", International Journal on Technical and Physical Problems of Engineering (IJTPE), Issue 55, Vol. 15, No. 2, pp. 125-133, June 2023.
[14] H.S. Rahli, N. Benamrane, "Intelligent Breast Cancer Screening Based on Deep Neural Networks", International Journal on Technical and Physical Problems of Engineering (IJTPE), Issue 57, Vol. 15, No. 4, pp. 404-409, December 2023.
[15] A. Gupta, V. Dengre, H.A. Kheruwala, M. Shah, "Comprehensive Review of Text-Mining Applications in Finance", Financ Innov, Vol. 6, p. 39, December 2020.
[16] J. Huang, R. Xing, Q. Li, "Asset Pricing Via Deep Graph Learning to Incorporate Heterogeneous Predictors", Int J. of Intelligent Sys., Vol. 37, pp. 8462-8489, November 2022.
[17] L. Zhu, H. Wu, M.T. Wells, "A News-Based Machine Learning Model for Adaptive Asset Pricing", Arxiv Preprint Arxiv:2106.07103, 2021.
[18] C. Zhang, "Asset Pricing and Deep Learning", arXiv:2209.12014, Vol. 24, [q-fin.ST], September 2022.
[19] R. Liu, F. Mai, Z. Shan, Y. Wu, "Predicting Shareholder Litigation on Insider Trading from Financial Text: An Interpretable Deep Learning Approach", Information and Management, Vol. 57, p. 103387, December 2020.
[20] S. Gholizadeh, N. Zhou, "Model Explainability in Deep Learning Based Natural Language Processing", arXiv:2106.07410 [cs], June 2021.
[21] L. Malandri, F.Z. Xing, C. Orsenigo, C. Vercellis, E. Cambria, "Public Mood-Driven Asset Allocation: The Importance of Financial Sentiment in Portfolio Management", Cong Compute, Vol. 10, pp. 1167-1176, December 2018.
[22] C.K. Soo, "Quantifying Sentiment with News Media Across Local Housing Markets", The Review of Financial Studies, Vol. 31, pp. 3689-3719, October 2018.
[23] S.B. Jabeur, C. Gharib, S. Mefteh Wali, W.B. Arfi, "CatBoost Model and Artificial Intelligence Techniques for Corporate Failure Prediction", Technological Forecasting and Social Change, Vol. 166, pp. 120-658, May 2021.
[24] J.Z.G. Hiew, X. Huang, H. Mou, D. Li, Q. Wu, Y. Xu, "BERT-Based Financial Sentiment Index and LSTM-Based Stock Return Predictability", Arxiv:1906.09024 [q-fin.ST], July 2022.
[25] M. Sedinkina, N. Breitkopf, H. Schutze, "Automatic Domain Adaptation Outperforms Manual Domain Adaptation for Predicting Financial Outcomes", The 57th

Annual Meeting of the Association for Computational Linguistics, pp. 346-359, 2019.

[26] B. Fazlija, P. Harder, "Using Financial News Sentiment for Stock Price Direction Prediction", Mathematics, Vol. 10, pp. 21-56, June 2022.

[27] A. Zaremba, E. Demir, "ChatGPT: Unlocking the Future of NLP in Finance", SSRN Journal, Modern Finance, 2023, Vol. 1, No. 1, pp. 93-98, 2023.

[28] L. Malandri, F.Z. Xing, C. Orsenigo, C. Vercellis, E. Cambria, "Public Mood–Driven Asset Allocation: The Importance of Financial Sentiment in Portfolio Management", Cong Compute, Vol. 10, pp. 1167-1176, December 2018.

[29] H. Peng, Y. Lin, M. Wu, "Bank Financial Risk Prediction Model Based on Big Data", Scientific Programming, Vol. 20, No. 22, pp. 1-9, February 2022.

[30] B. Sandwidi, S. Pallitharammal Mukkolakal, "Transformers-Based Approach for a Sustainability Term-Based Sentiment Analysis (STBSA)", The Second Workshop on NLP for Positive Impact (NLP4PI), (Abu Dhabi, United Arab Emirates (Hybrid)), pp. 157-170, Association for Computational Linguistics, 2022.

## BIOGRAPHIES

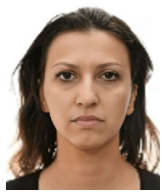

Name: **Denisa**
Surname: **Millo**
Birthday: 19.11.1990
Birthplace: Elbasan, Albania
Bachelor: Business Informatics, Department of Statistics and Applied Informatics, Faculty of Economy, University of Tirana, Tirana, Albania, 2012
Master: Operational Research in Management, Department of Statistics and Applied Informatics, Faculty of Economy, University of Tirana, Tirana, Albania, 2014
Doctorate: Student, Information Systems in Economy, Department of Statistics and Applied Informatics, Faculty of Economy, University of Tirana, Tirana, Albania, Since 2022
The Last Scientific Position: Lecturer, Department of Statistics and Applied Informatics, Faculty of Economy, University of Tirana, Tirana, Albania, Since 2016
Research Interests: Natural Language Processing in Finance and Media
Scientific Publications: 8 Papers, 2 Projects
Scientific Memberships: UNINOVIS Alumni

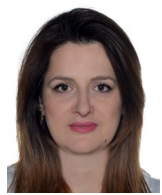

Name: **Blerina**
Surname: **Vika**
Birthday: 27.06.1984
Birthplace: Tirana, Albania
Education: Informatics, Department of Informatics, Faculty of Natural Sciences, University of Tirana, Tirana, Albania, 2007
Master: Advanced Informatics, Department of Informatics, Faculty of Natural Sciences, University of Tirana, Tirana, Albania, 2011
Doctorate: Information Systems, Department of Statistics and Applied Informatics, Faculty of Natural Sciences, University of Tirana, Tirana, Albania, 2010
The Last Scientific Position: Assoc. Prof., Department of Statistics and Applied Informatics, Faculty of Economy, University of Tirana, Tirana, Albania, 2024
Research Interests: Artificial Intelligences Machine Learning, Forecasting Models
Scientific Publications: 20 Papers, 2 Books, 4 Projects

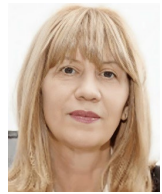

Name: **Nevila**
Surname: **Baci**
Birthday: 15.05.1965
Birthplace: Tirana, Albania
Bachelor: Economics, Department of Economics, Faculty of Economy, University of Tirana, Tirana, Albania, 1987
Doctorate: E-government, Faculty of Economy, University of Tirana, Tirana, Albania, 2005
The Last Scientific Position: Prof., Department of Statistics and Applied Informatics, Faculty of Economy, University of Tirana, Tirana, Albania, 2022
Research Interests: Informatics, Information Systems, E-Commerce
Scientific Publications: 25 Papers, 1 Books, 5 Projects
Scientific Memberships: Academy of Science, Albania